\documentclass[10pt,twocolumn,letterpaper]{article}

\usepackage{times}
\usepackage{epsfig}
\usepackage{graphicx}
\usepackage{amsmath}
\usepackage{amssymb}
\usepackage{multirow}
\DeclareMathOperator*{\argmax}{argmax}

\usepackage[pagebackref=true,breaklinks=true,letterpaper=true,colorlinks,bookmarks=false]{hyperref}

\usepackage{xspace}

\makeatletter
\DeclareRobustCommand\onedot{\futurelet\@let@token\@onedot}
\def\@onedot{\ifx\@let@token.\else.\null\fi\xspace}
\def\eg{\emph{e.g}\onedot} 
\def\ie{\emph{i.e}\onedot}

\def\etal{\emph{et al}\onedot}

\begin{document}

\title{Part-Stacked CNN for Fine-Grained Visual Categorization}

\author{Shaoli Huang*\\
University of Technology, Sydney\\
Sydney, Ultimo, NSW 2007, Australia\\
{\tt\small shaoli.huang@student.uts.edu.au}
\and
Zhe Xu*\\
Shanghai Jiao Tong University\\
Shanghai, 200240, China\\
{\tt\small xz3030@sjtu.edu.cn}
\and
Dacheng Tao\\
University of Technology, Sydney\\
Sydney, Ultimo, NSW 2007, Australia\\
{\tt\small dacheng.tao@uts.edu.au}
\and
Ya Zhang\\
Shanghai Jiao Tong University\\
Shanghai, 200240, China\\
{\tt\small ya\_zhang@sjtu.edu.cn}
}

\maketitle

\begin{abstract}
   
   In the context of fine-grained visual categorization, the ability to interpret models as human-understandable visual manuals is sometimes as important as achieving high classification accuracy. In this paper, we propose a novel Part-Stacked CNN architecture that explicitly explains the fine-grained recognition process by modeling subtle differences from object parts. Based on manually-labeled strong part annotations, the proposed architecture consists of a fully convolutional network to locate multiple object parts and a two-stream classification network that encodes object-level and part-level cues simultaneously. By adopting a set of sharing strategies between the computation of multiple object parts, the proposed architecture is very efficient running at $20$ frames/sec during inference. Experimental results on the CUB-200-2011 dataset reveal the effectiveness of the proposed architecture, from both the perspective of classification accuracy and model interpretability.
\end{abstract}


\section{Introduction}
\label{sec:intro}
Fine-grained visual categorization aims to distinguish objects at the subordinate level, \eg different species of birds \cite{welinder2010caltech,wah2011caltech,berg2014birdsnap}, pets \cite{khosla2011novel,parkhi2012cats}, flowers \cite{nilsback2008automated,angelova2013image} and cars \cite{stark2011fine,maji2013fine}. It is a highly challenging task due to the small inter-class variance caused by highly similar subordinate categories, and the large intra-class variance by nuisance factors such as pose, viewpoint and occlusion. Inspiringly, huge progress has been made over the last few years \cite{wah2011multiclass,berg2014birdsnap,vedaldi2014understanding,krause2015fine,xu2015augmenting}, making fine-grained recognition techniques a large step closer to practical use in various applications, such as wildlife observation and surveillance systems.


\begin{figure}[t]
\begin{center}
\includegraphics[width=.9\linewidth]{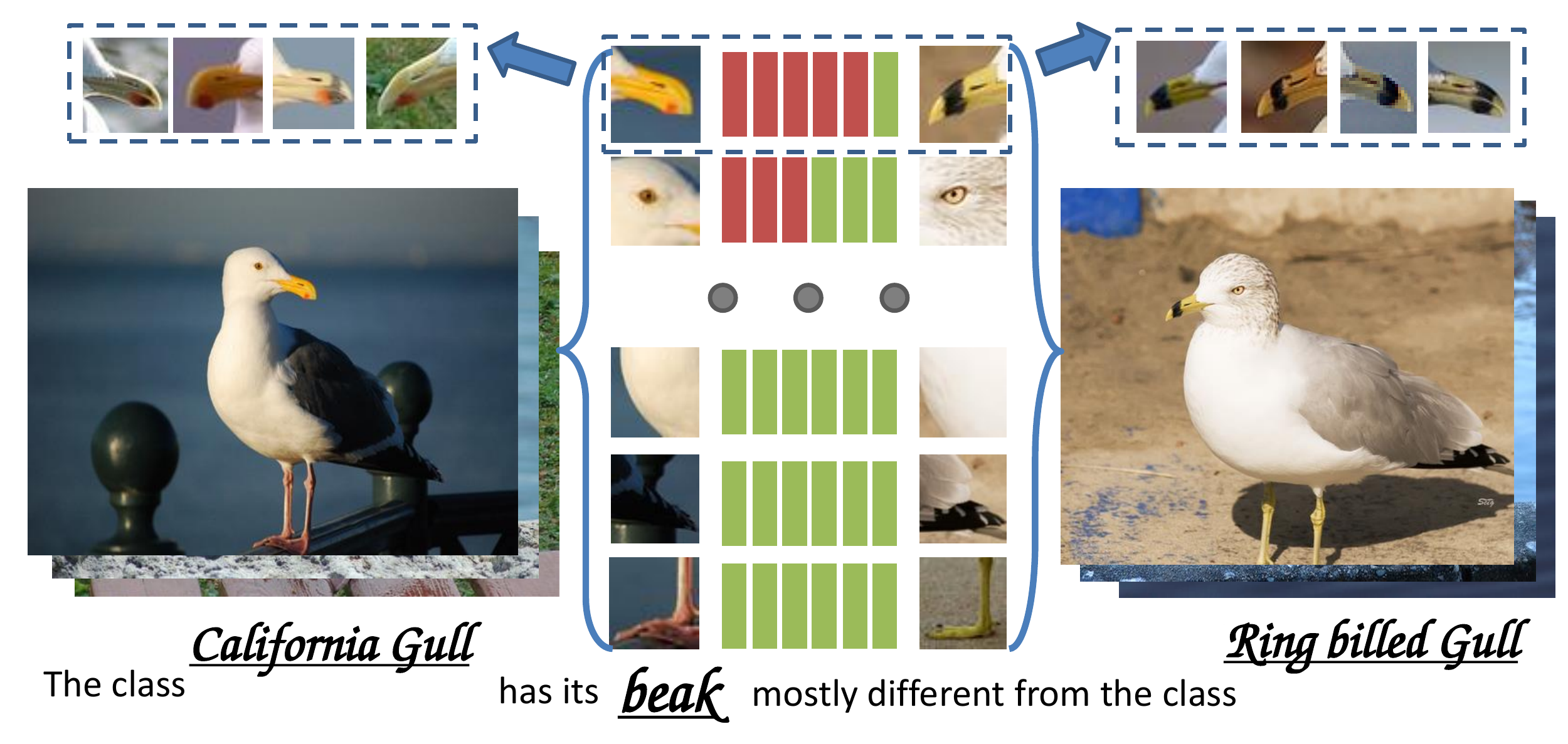}
\end{center}
   \caption{Overview of the proposed approach. We propose to classify fine-grained categories by modeling the subtle difference from specific object parts. Beyond classification results, the proposed PS-CNN architecture also offers human-understandable instructions on how to classify highly similar object categories explicitly.}
\label{fig:title}
\end{figure}

Whilst numerous attempts have been made to boost the classification accuracy of fine-grained visual categorization \cite{deng2013fine,chai2013symbiotic,branson2014bird,lin2015bilinear}, we argue that another important aspect of the problem has yet been severely overlooked, \ie, the ability to generate a human-understandable ``manual'' on how to distinguish fine-grained categories in detail. For example, volunteers for ecological protection may certainly benefit from an algorithm that could not only classify bird species accurately, but also provide brief instructions on how to distinguish a category from its most similar subspecies - \eg, a salient difference between a \textit{Ringed-billed gull} and a \textit{California gull} lies in the pattern on their beaks (Figure \ref{fig:title}) - with some intuitive illustration examples. Existing fine-grained recognition methods that aim to provide a visual field guide mostly follow the routine of ``part-based one-vs-one features'' (POOFs) \cite{berg2013poof,berg2013you,berg2014birdsnap} or employ human-in-the-loop methods \cite{kumar2012leafsnap,branson2014ignorant,van2015building}.
Since the data size has been increasing drastically, a method that simultaneously implements and interprets fine-grained visual categorization using the latest deep learning methods \cite{krizhevsky2012imagenet} is therefore highly advocated.


It is widely acknowledged that the subtle difference between fine-grained categories mostly resides in the unique properties of object parts \cite{rosch1976basic,berg2013poof,chai2013symbiotic,maji2014part,zhang2014part}. Therefore, a practical solution to interpret classification results as human-understandable manuals is to discover classification criteria from object parts. Some of existing fine-grained datasets have provided detailed part annotations including part landmarks and attributes \cite{wah2011caltech,maji2013fine}. However, they are usually associated with a large number of object parts, which poses heavy computational burden for both part detection and classification. From this perspective, one would like to seek a method that follows the object-part-aware strategy to provide interpretable predicting criteria, while requiring minimum computational effort to deal with a possibly large number of parts.


In this paper, we propose a new part-based CNN architecture for fine-grained visual categorization that models multiple object parts in a unified framework with high efficiency.
Similar with previous fine-grained recognition approaches, the proposed method consists of a localization module to detect object parts (``where pathway'') and a classification module to classify fine-grained categories at the subordinate level (``what pathway''). In particular, we employ a fully convolutional network (FCN) to perform object part localization. The inferred part locations are fed into the classification network, in which a two-stream architecture is proposed to analyze images in both object-level (bounding boxes) and part-level (part landmarks).
The computation of multiple parts is first conducted via a shared feature extraction route, then separated through a novel part crop layer, concatenated, and then fed into a shallower network to perform object classification.
Except for categorical predictions, the proposed method also generates interpretable classification instructions based on object parts.
Since the proposed architecture employs a sharing strategy that stacks the computation of multiple parts together, we call it \emph{Part-Stacked CNN} (PS-CNN).

The contributions of this paper include: 1) we present a novel and efficient part-based CNN architecture for fine-grained recognition; 2) our architecture adopts an FCN to localize object parts, which has seldom been studied before in the context of object recognition; 3) our classification network follows a two-stream structure that captures both object-level and part-level information, in which a new share-and-divide strategy is presented on the computation of multiple object parts. As a result, the proposed architecture is very efficient, with a capacity of $20$ frames/sec\footnote{For reference, a single CaffeNet runs at $50$ frames/sec under the same experimental setting.} on a Tesla K80 to classify images at test time using $15$ object parts; 4) to the best of our knowledge, the proposed method is the first attempt to both implement and explicitly interpret the process of fine-grained visual categorization using deep learning methods.
The effectiveness of the proposed method is demonstrated through systematic experiments on the Caltech-UCSD Birds-200-2011 \cite{wah2011caltech} dataset, in which we achieved $76\%$ classification accuracy. We also present practical examples of human-understanding manuals generated by the proposed method for the task of fine-grained visual categorization.


The rest of the paper is organized as follows. Section \ref{sec:relatedwork} summarizes related works. The proposed architecture including the localization network and the classification network is described in Section \ref{sec:pscnn}.
Detailed performance studies and analysis are conducted in Section \ref{sec:exp}. Section \ref{sec:conclusion} concludes the paper and proposes discussions on the application scenarios of the proposed PS-CNN.


\section{Related Work}\label{sec:relatedwork}
\noindent\textbf{Fine-Grained Visual Categorization}.
A number of methods have been developed to classify object categories at the subordinate level. Recently, the best performing methods mostly sought for improvement brought by the following three aspects: more discriminative features including deep CNNs for better visual representation \cite{bo2010kernel,sanchez2011fisher,krizhevsky2012imagenet,szegedy2014going,simonyan2014very}, explicit alignment approaches to eliminate pose displacements \cite{branson2014bird,gavves2014local}, and part-based methods to study the impact of object parts \cite{berg2013poof,zhang2014panda,maji2014part,zhang2014part,gkioxari2014actions}.
Another line of research explored human-in-the-loop methods \cite{branson2010visual,deng2013fine,wah2014similarity} to identify the most discriminative regions for classifying fine-grained categories. Although such methods provided direct references of how people perform fine-grained recognition in real life, they were impossible to scale for large systems due to the need of human interactions at test time.



Current state-of-the-art methods for fine-grained recognition are part-based R-CNN by Zhang \etal \cite{zhang2014part} and Bilinear CNN by Lin \etal \cite{lin2015bilinear}, which both employed a two-stage pipeline of part detection and part-based object classification. The main idea of the proposed PS-CNN is largely inherited from \cite{zhang2014part}, who first detected the location of two object parts and then trained an individual CNN based on the unique properties of each part. Compared to part-based R-CNN, the proposed method is far more efficient in both detection and classification phrases. As a result, we are able to employ much more object parts than that of \cite{zhang2014part}, while still being significantly faster at test time.

\begin{figure*}[Ht]
\begin{center}
\includegraphics[width=.9\linewidth]{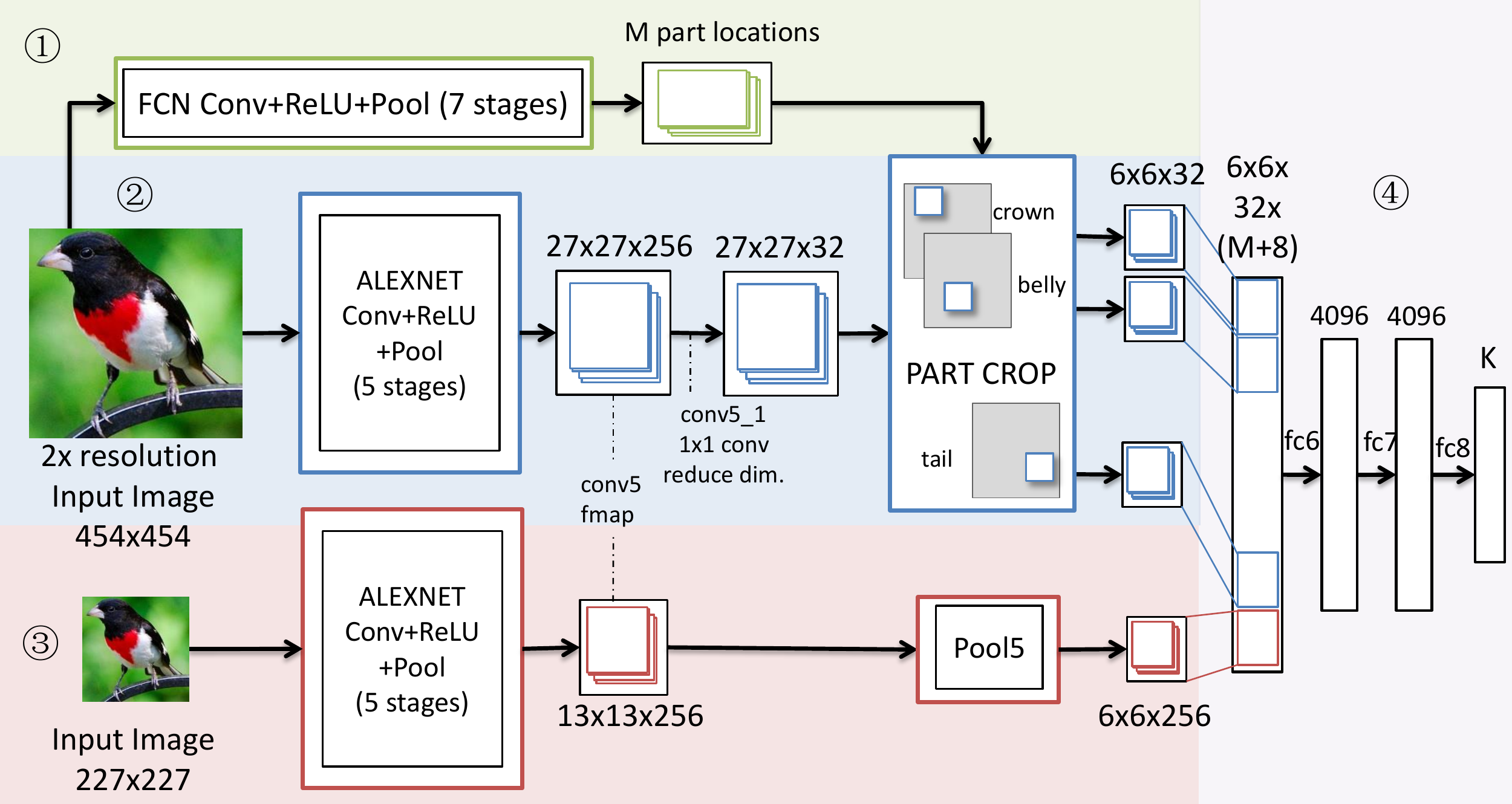}
\end{center}
   \caption{Network architecture of the proposed Part-Stacked CNN model. The model consists of: 1) a fully convolutional network for part landmark localization; 2) a part stream where multiple parts share the same feature extraction procedure, while being separated by a novel part crop layer given detected part locations; 3) an object stream with lower spatial-resolution input images to capture bounding-box level supervision; and 4) three fully connected layers to achieve the final classification results based on a concatenated feature map containing information from all parts and the bounding box.}
\label{fig:architecture}
\end{figure*}

\begin{figure*}[t]
\begin{center}
\includegraphics[width=.9\linewidth]{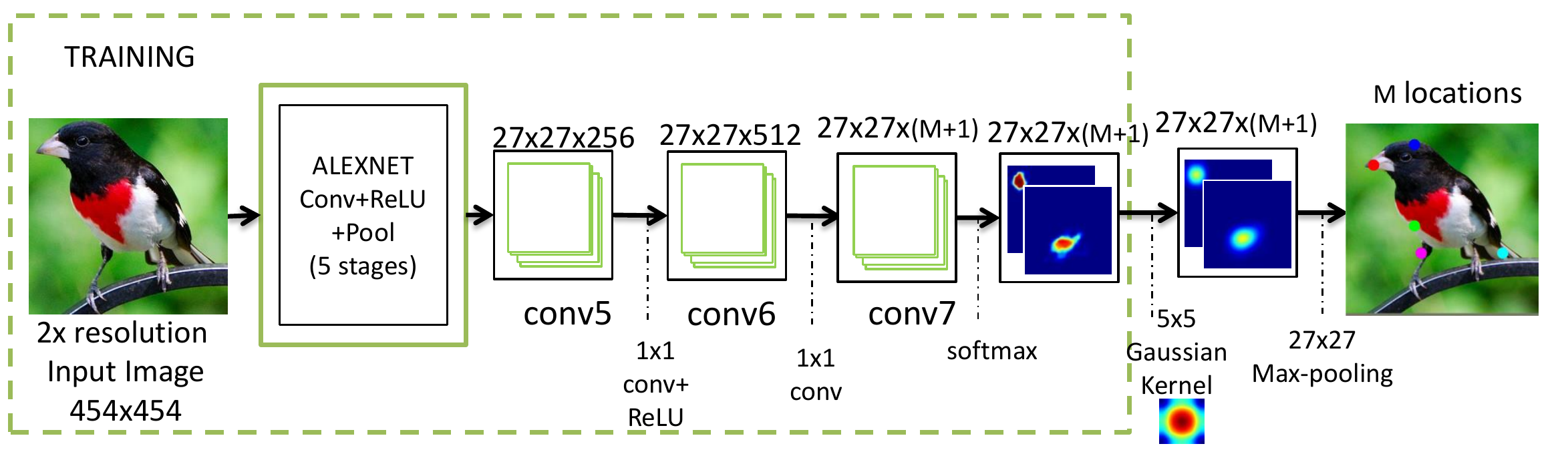}
\end{center}
   \caption{Demonstration of the localization network. Training process is denoted inside the dashed box. For inference, a Gaussian kernel is then introduced to remove noise. The results are $M$ 2D part locations in the $27\times27$ \textit{conv5} feature map.}
\label{fig:fcn}
\end{figure*}


On the other hand, Lin \etal \cite{lin2015bilinear} argued that manually defined parts were sub-optimal for the task of object recognition, and thus proposed a bilinear model consisting of two streams whose roles were interchangeable as detectors or features. Although this design enjoyed the data-driven nature that could possibly lead to optimal classification performance, it also made the resultant model hard to interpret. On the contrary, our method tries to balance the need of both both classification accuracy and model interpretability in fine-grained recognition systems. \\





\noindent\textbf{Fully Convolutional Networks}.
Fully convolutional network (FCN) is a fast and effective approach to produce dense prediction with convolutional networks. Successful examples can be found on tasks including sliding window detection \cite{sermanet2013overfeat}, semantic segmentation \cite{long2015fully}, and human pose estimation \cite{tompson2014joint}. We find the problem of part landmark localization in fine-grained recognition closely related to human pose estimation, in which a critical step is to detect a set of key points indicating multiple components of human body.


\section{Part-Stacked CNN} \label{sec:pscnn}

We present the model architecture of the proposed Part-Stacked CNN in this section. In accordance with the common framework for fine-grained recognition, the proposed architecture is decomposed into a \emph{Localization Network} (Section \ref{subsec:localization}) and a \emph{Classification Network} (Section \ref{subsec:classification}). 
We adopt CaffeNet \cite{jia2014caffe}, a slightly modified version of the standard seven-layer AlexNet \cite{krizhevsky2012imagenet} architecture, as the basic structure of the network; deeper networks could potentially lead to better recognition accuracy, but may also result in lower efficiency. 

A unique design in our architecture is that the message transferring operation from the localization network to the classification network, \ie using detected part locations to perform part-based classification, is conducted directly on the \textit{conv5} output feature maps within the process of data forwarding. It is a significant difference compared to the standard two-stage pipeline of part-based R-CNN \cite{zhang2014part} that consecutively localizes object parts and then trains part-specific CNNs on the detected regions. Based on this design, a set of sharing schemes are performed to make the proposed PS-CNN fairly efficient for both learning and inference. Figure \ref{fig:architecture} illustrates the overall network architecture.



\subsection{Localization Network} \label{subsec:localization}
The first stage of the proposed architecture is a localization network that aims to detect the location of object parts. We employ the simplest form of part landmark annotations, \ie a 2D key point is annotated at the center of each object part. Assume that $M$ - the number of object parts labeled in the dataset, is sufficient large to offer a complete set of object parts on which fine-grained categories are usually different from each other.
Motivated by recent progress of human pose estimation \cite{long2015fully} and semantic segmentation \cite{tompson2014joint}, we adopt a fully convolutional network (FCN) \cite{matan1995multi} to generate dense output feature maps for locating object parts.\\


\noindent\textbf{Fully convolutional network.}
A fully convolutional network is achieved by replacing the parameter-rich fully connected layers in standard CNN architectures by convolutional layers with $1\times1$ kernels. Given an input RGB image, the output of a fully convolutional network is a \textit{feature map} in reduced dimension compared to the input. The computation of each unit in the feature map only corresponds to pixels inside a region with fixed size in the input image, which is called its \textit{receptive field}.
FCN is preferred in our framework due to the following three reasons: 1) feature maps generated by FCN can be directly utilized as the part locating results in the classification network, which will be detailed in Section \ref{subsec:classification}; 2) results of multiple object parts can be obtained simultaneously using an FCN; 3) FCN is very efficient in both learning and inference. \\


\noindent\textbf{Learning.} We model the part localization process as a multi-class classification problem on dense output spatial positions. In particular, suppose the output of the last convolutional layer in the FCN is in the size of $h\times w\times d$, where $h$ and $w$ are spatial dimensions and $d$ is the number of channels. We set $d=M+1$. Here $M$ is the number of object parts and $1$ denotes for an additional channel to model the background. To generate corresponding ground-truth labels in the form of feature maps, units indexed by $h \times w$ spatial positions are labeled by their nearest object part; units that are not close to any of the labeled parts (with an overlap $<0.5$ with respect to receptive field) are labeled as background.


A practical problem here is to determine the model depth and the size of input images for training the FCN. Generally speaking, layers at later stages carry more discriminative power and thus are more likely to generate promising localization results; however, their receptive fields are also much larger than those of previous layers. For example, the receptive field of \textit{conv5} layer in CaffeNet has a size of $163\times163$ compared to the $227\times227$ input image, which is too large to model an object part. We propose a simple trick to deal with this problem, \ie, upsampling the input images so that the fixed-size receptive fields denoting object parts become relatively smaller compared to the whole object, while still being able to use layers at later stages to guarantee enough discriminative power.


The localization network in the proposed PS-CNN is illustrated in Figure \ref{fig:fcn}. The input of the FCN is a bounding-box-cropped RGB image, warped and resized into a fixed size of $454\times454$. The structure of the first five layers is identical to those in CaffeNet, which leads to a $27\times27\times256$ output after \textit{conv5} layer. Afterwards, we further introduce a $1\times1$ convolutional layer with $512$ output channels as \textit{conv6}, and another $1\times1$ convolutional layer with $M+1$ outputs termed \textit{conv7} to perform classification. By adopting a spatial preserving softmax that normalizes predictions at each spatial location of the feature map, the final loss function is a sum of softmax loss at all $27\times27$ positions:

\begin{equation}\label{eqn:fcnloss}
L = -\sum_{h=1}^{27}\sum_{w=1}^{27} \log\sigma(h,w,\hat{c}),
\end{equation}
where
\begin{displaymath}
\sigma(h,w,\hat{c}) = \frac{\exp(f_{conv7}(h,w,\hat{c}))}{\sum_{c=0}^M \exp(f_{conv7}(h,w,c))}.
\end{displaymath}
Here, $\hat{c}\in[0,1,...,M]$ is the part label of the patch at location $(h,w)$, where the label $0$ denotes background. $f_{conv7}(h,w,c)$ stands for the output of \textit{conv7} layer at spatial position $(h,w)$ and channel $c$.
\\

\noindent\textbf{Inference.} The inference process starts from the output of the learned FCN, \ie, $(M+1)$ part-specific heat maps in the size of $27\times27$, in which we introduce a Gaussian kernel $\mathcal{G}$ to remove isolated noise in the feature maps. The final output of the localization network are $M$ locations in the $27\times27$ \textit{conv5} feature map, each of which is computed as the location with the maximum response for one object part. 

Meanwhile, considering that object parts may be missing in some images due to varied poses and occlusion, we set a threshold $\mu$ that if the maximum response of a part is below $\mu$, we simply discard this part's channel in the classification network for this image. Let $g(h,w,c)=\sigma(h,w,c)*\mathcal{G}$, the inferred part locations are given as:

\begin{equation}
(h_c^*,w_c^*)=
  \left\{
   \begin{array}{ll}
   \argmax_{h,w} g(h,w,c) & \text{if } g(h_c^*,w_c^*,c)>\mu, \\
   (-1,-1) & \text{otherwise.}
   \end{array}
  \right.
\end{equation}

\begin{figure}[t]
\begin{center}
\includegraphics[width=.9\linewidth]{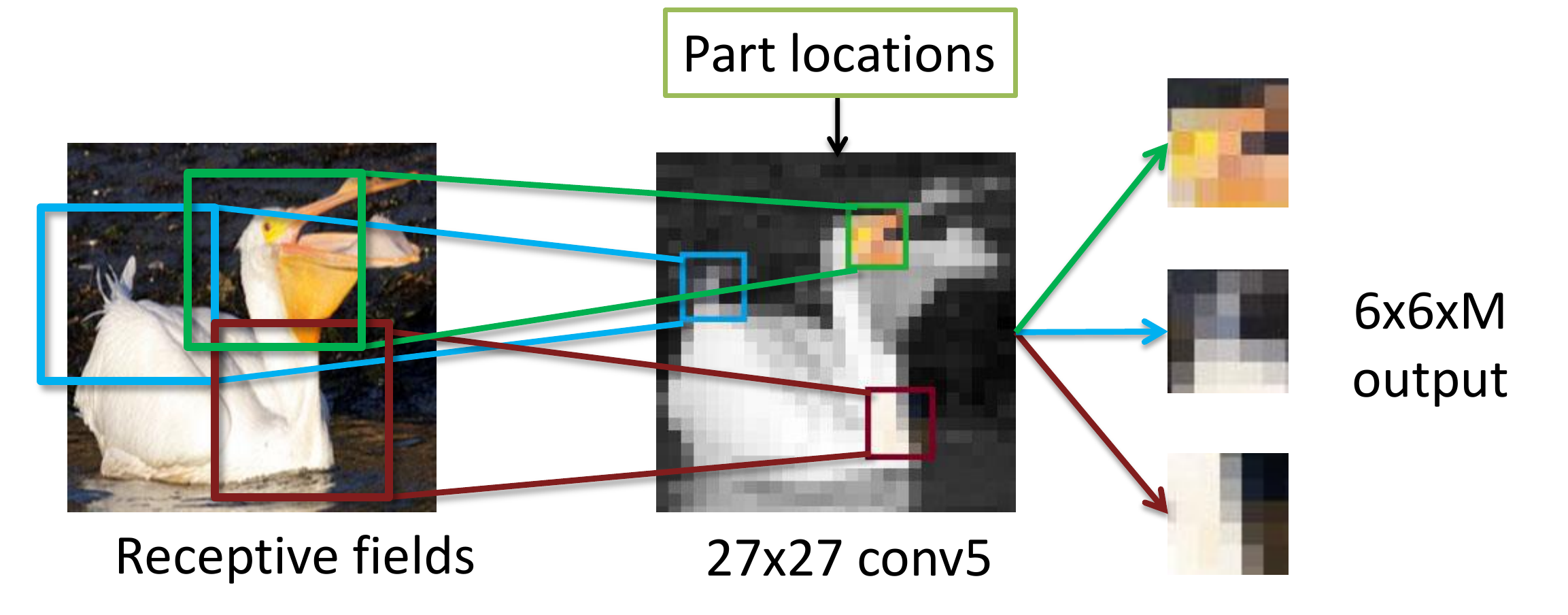}
\end{center}
   \caption{Demonstration of the part crop layer with three object parts. The input is a $27\times27$ heatmap and three object parts' anchor position, and the output is three $6\times6$ regions centered at the respective anchor position. Image in the left shows the respective field for each part after performing part cropping.}
\label{fig:partcrop}
\end{figure}

\subsection{Classification network}\label{subsec:classification}
The second stage of the proposed PS-CNN is a classification network with the inferred part locations given as an input. It follows a two-stream architecture with a \emph{Part Stream} and a \emph{Object Stream} to capture semantics from multiple levels. A sub-network consisting of three fully connected layers is then performed as an object classifier, as shown in Figure \ref{fig:architecture}. \\

\noindent\textbf{Part stream.}
The part stream acts as the core of the proposed PS-CNN architecture. To capture object-part-dependent differences between fine-grained categories, one can train a set of part CNNs, each one of which conducts classification on a part separately, as proposed by Zhang \etal \cite{zhang2014part}. Although such method worked well for \cite{zhang2014part} who only employed two object parts, we argue that it is not applicable when the number of object parts is much larger in our case, because of the high time and space complexity.


In PS-CNN, we introduce two strategies to improve the efficiency of the part stream. The first one is model parameter sharing. Specifically, model parameters of the first five convolutional layers are shared among all object parts, which can be regarded as a generic part-level feature extractor. This strategy leads to less parameters in the proposed architecture and thus reduces the risk of overfitting.




Other than model parameter sharing, we also conduct a computational sharing strategy. The goal is to make sure that the feature extraction procedure of all parts only requires one pass through the convolutional layers. Analogous to the localization network, the input images of the part stream are in doubled resolution $454\times454$ so that the respective receptive fields are not too large to model object parts; forwarding the network to \textit{conv5} layer generates output feature maps of size $27\times 27$. By far, the computation of all object parts is completely shared.

After performing the shared feature extraction procedure, the computation of each object part is then partitioned through a \emph{part crop layer} to model part-specific classification cues. As shown in Figure \ref{fig:partcrop}, the input for the part crop layer is a set of feature maps, \eg, the output of \textit{conv5} layer in our architecture, and the predicted part locations from the previous localization network, which also reside in \textit{conv5} feature maps. For each part, the part crop layer extracts a local neighborhood region centered at the detected part location. Features outside the cropped region are simply dropped. In practice, we crop $6\times 6$ neighborhood regions out of the $27\times 27$ \textit{conv5} feature maps to match the output size of the object stream. The resultant receptive fields for the cropped feature maps has a width of $163+16\times 5=243$.\\


%

\noindent\textbf{Object stream.}
The object stream utilizes bounding-box-level supervision to capture object-level semantics for fine-grained recognition. It follows the general architecture of CaffeNet, in which the input of the network is a $227\times 227$ RGB image and the output of \textit{pool5} layer are $6\times 6$ feature maps.


We find the design of the two-stream architecture in PS-CNN analogous to the famous Deformable Part-based Models \cite{felzenszwalb2010object}, in which object-level features are captured through a root filter in a coarser scale, while detailed part-level information is modeled by several part filters at a finer scale. We find it critical to measure visual cues from multiple semantic levels in an object recognition algorithm.\\

\noindent\textbf{Dimension reduction and fully connected layers.}
The aforementioned two-stream architecture generates an individual feature map for each object part and bounding box. When conducting classification, they serve as an over-complete set of CNN features from multiple scales. Following the standard CaffeNet architecture, we employ a DNN including three fully connected layers as object classifiers. The first fully connected layer \emph{fc6} now becomes a part concatenation layer whose input is generated by stacking the output feature maps of the part stream and the object stream together. However, such a concatenating process requires $M+1$ times more model parameters than the original \textit{fc6} layer in CaffeNet, which leads to a huge memory cost.


To reduce model parameters, we introduce a $1\times1$ convolutional layer termed \textit{conv5\_1} in the part stream that projects the $256$ dimensional \textit{conv5} output to $32$-d. It is identical to a low-rank projection of the model output and thus can be initialized through standard PCA. Nevertheless, in our experiments, we find that directly initializing the weights of the additional convolution by PCA in practice worsens the performance. To enable domain-specific fine-tuning from pre-trained CNN model weights, we train an auxiliary CNN to initialize the weights for the additional convolutional layer.

Let $X^c\in \mathbb{R}^{N\times M\times 6\times6}$ be the $c^{th}$ $6\times6$ cropped region around the center point $(h^*_c,w^*_c)$ from \textit{conv5\_1} feature maps $X \in \mathbb{R}^{N\times M\times27\times27}$, where $(h^*_c, w^*_c)$ is the predicted location for part $c$. The output of part concatenation layer \emph{fc6} can be formulated as:
\begin{equation}
f_{out}(X) = \sigma(\sum_{c=1}^{M} (W^c)^TX^c),
\end{equation}
where $W^c$ is the model parameters for part $c$ in \emph{fc6} layer, and $\sigma$ is an activation function.

We conduct the standard gradient descent method to train the classification network. The most complicated part for computing gradients lies in the dimension reduction layer due to the impact of part cropping. Specifically, the gradient of each cropped part feature map (in $6\times 6$ spatial resolution) is projected back to the original size of \textit{conv5} ($27\times27$ feature maps) according to the respective part location and then summed up.
Note that the proposed PS-CNN is implemented as a two stage framework, \ie after training the FCN, weights of the localization network are fixed when training the classification network.



\section{Experiments}\label{sec:exp}
We present experimental results and analysis of the proposed method in this section. Specifically, we will evaluate the performance through four different aspects: localization accuracy, classification accuracy, inference efficiency, and model interpretation.

\subsection{Dataset and implementation details}
Experiments are conducted on the widely used fine-grained classification benchmark the Caltech-UCSD Birds dataset (CUB-200-2011) \cite{wah2011caltech}. The dataset contains $200$ bird categories with roughly $30$ training images per category. In the training phase we adopt strong supervision available in the dataset, \ie we employ 2D key point part annotations of altogether $M=15$ object parts together with image-level labels and object bounding boxes.



The proposed Part-Stacked CNN architecture is implemented using the open-source package Caffe \cite{jia2014caffe}.
Specifically, bounding-box cropped input images are warped to a fixed size of $512\times512$, randomly cropped into $454\times454$, and then fed into the localization network and the part stream in the classification network as input. We employ a pooling layer in the object stream that downsamples the $454\times454$ input to $227\times227$ to guarantee synchronization between the two streams in the classification network.

\begin{table}
\begin{center}
\small
\begin{tabular}{l|c|c|c}
\hline
Model architecture & MPK & MRK & APK \\
\hline
conv5+cls & 70.0 & 80.6 & 83.5 \\
conv5+conv6(256)+cls & 71.3 & 81.8 & 84.7 \\
conv5+conv6(512)+cls & 71.5 & 81.9 & 84.8 \\
conv5+conv6(512)+cls+gaussian & 80.0 & 83.8 & 86.6 \\
\hline
\end{tabular}
\end{center}
\caption{Comparison of different model architectures on localization results. ``conv5'' stands for the first 5 convolutional layers in CaffeNet; ``conv6(256)'' stands for the additional $1\times1$ convolutional layer with 256 output channels; ``cls'' denotes the classification layer with $M+1$ output channels; ``gaussian'' represents a Gaussian kernel for smoothing.}
\label{tab:locarc}
\end{table}

\begin{table*}[ht]
\small
\begin{center}
\begin{tabular}{l|c|c|c|c|c|c|c|c}
\hline
part & throat & beak & crown & forehead & right eye & nape & left eye & back\\
\hline
APK & 0.908 & 0.894 & 0.894 & 0.885 & 0.861 & 0.857 & 0.850 & 0.807\\
\hline\hline
part & breast & belly & right leg & tail & left leg & right wing & left wing & overall\\
\hline
APK & 0.799 & 0.794 & 0.775 & 0.760 & 0.750 & 0.678 & 0.670 & 0.866\\
\hline
\end{tabular}
\end{center}
\caption{\emph{APK} for each object part in the CUB-200-2011 test set in descending order.}
\label{tab:locapk}
\end{table*}

\begin{figure*}[t]
\begin{center}
\includegraphics[width=.88\linewidth]{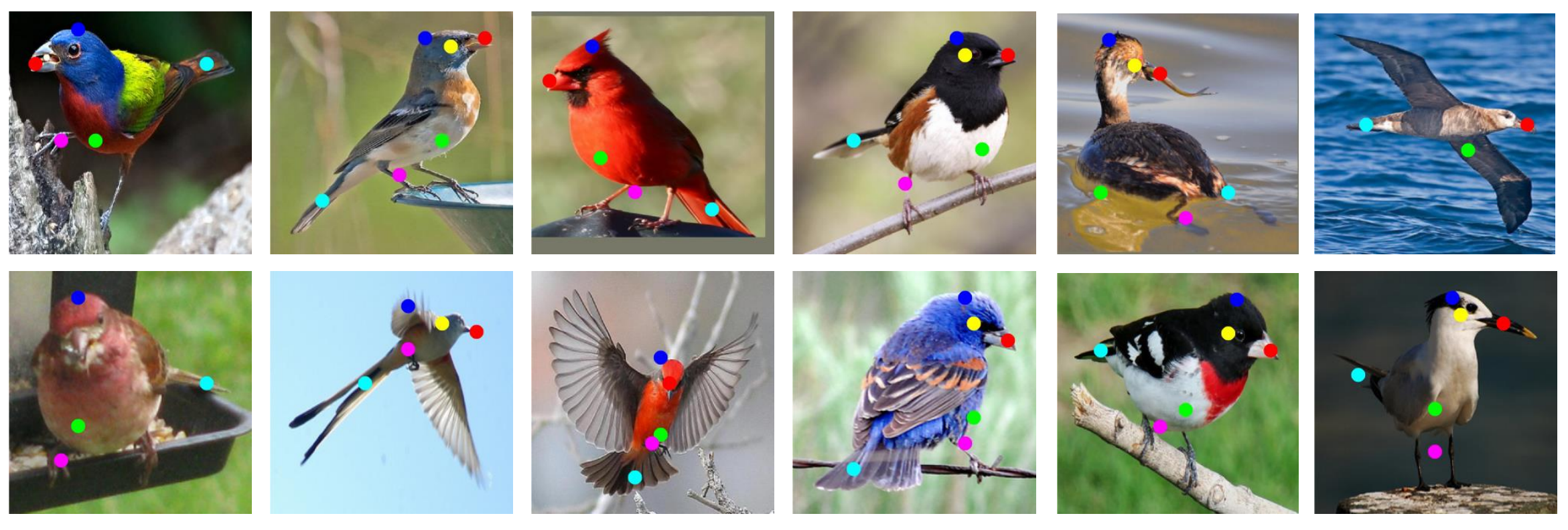}
\end{center}
   \caption{Typical localization results on CUB-200-2011 test set. We show 6 of the 15 detected parts here. They are: beak (red), belly (green), crown (blue), right eye (yellow), right leg (magenta), tail (cyan). Better viewed in color.}
\label{fig:loc}
\end{figure*}

\subsection{Localization results}
We quantitatively assess the localization correctness using three metrics. The first two are \emph{MPK} (Mean Precision of Key points over images) and \emph{MRK} (Mean Recall of Key points over images), which calculate precision and recall for the detected key points in each image and then average them over all test images. Suppose that image $I$ has $n_{gt}$ ground-truth parts, the proposed method predicts $n_{pd}$ parts where $n_{tp}$ of them is correctly located, \emph{MPK} and \emph{MRK} are computed as:

\begin{eqnarray}
\text{MPK} = \frac{1}{N}\sum_{I\in\mathcal{I}}\frac{n_{tp}}{n_{pd}} \text{,    MRK} = \frac{1}{N}\sum_{I\in\mathcal{I}}\frac{n_{tp}}{n_{gt}},
\end{eqnarray}
where $N$ is the number of test images in the dataset.

We also adopt \emph{APK} (Average Precision of Key points) \cite{yang2013articulated} to explicitly study the localization performance for each part.
Following \cite{long2014convnets}, we consider a key point to be correctly predicted if the prediction lies within a Euclidean distance of $\alpha$ times
the maximum of the bounding box width and height compared to the ground truth.
We set $\alpha=0.1$ in all the analysis below.

The results of different FCN architectures evaluated in this paper are outlined in Table \ref{tab:locarc}. By introducing an additional $1\times1$ convolutional layer after the first five layers in CaffeNet, a reasonably significant performance improvement is achieved. The Gaussian kernel also contributes to the localization accuracy by removing isolated noise, achieving a nearly $10\%$ improvement in \emph{MPK} in particular. The final localization network achieves an inspiring $86.6\%$ \emph{APK} on the test set of CUB-200-2011 for 15 object parts.

\subsection{Classification results}
\begin{table}[t]
\small
\begin{center}
\begin{tabular}{c|c|c|c|c}
\hline
BBox only & +2 part & +4 part & +8 part & +15 part\\
\hline
69.08 & 73.72 & 74.84 & 76.22 & 76.15\\
\hline
\end{tabular}
\end{center}
\caption{The effect of increasing the number of object parts on the classification accuracy.}
\label{tab:increm}
\end{table}

Furthermore, we present per part \emph{APK}s in Table \ref{tab:locapk}. An interesting phenomenon here is that parts residing near the head of the birds tend to be located more accurately. It turns out that the birds' head has relatively more stable structure with less deformations and lower probability to be occluded. On the contrary, parts that are highly deformable such as wings and legs get lower \emph{APK} values. Figure \ref{fig:loc} shows typical localization results of the proposed method.

\begin{figure*}[t]
\begin{center}
\includegraphics[width=.9\linewidth]{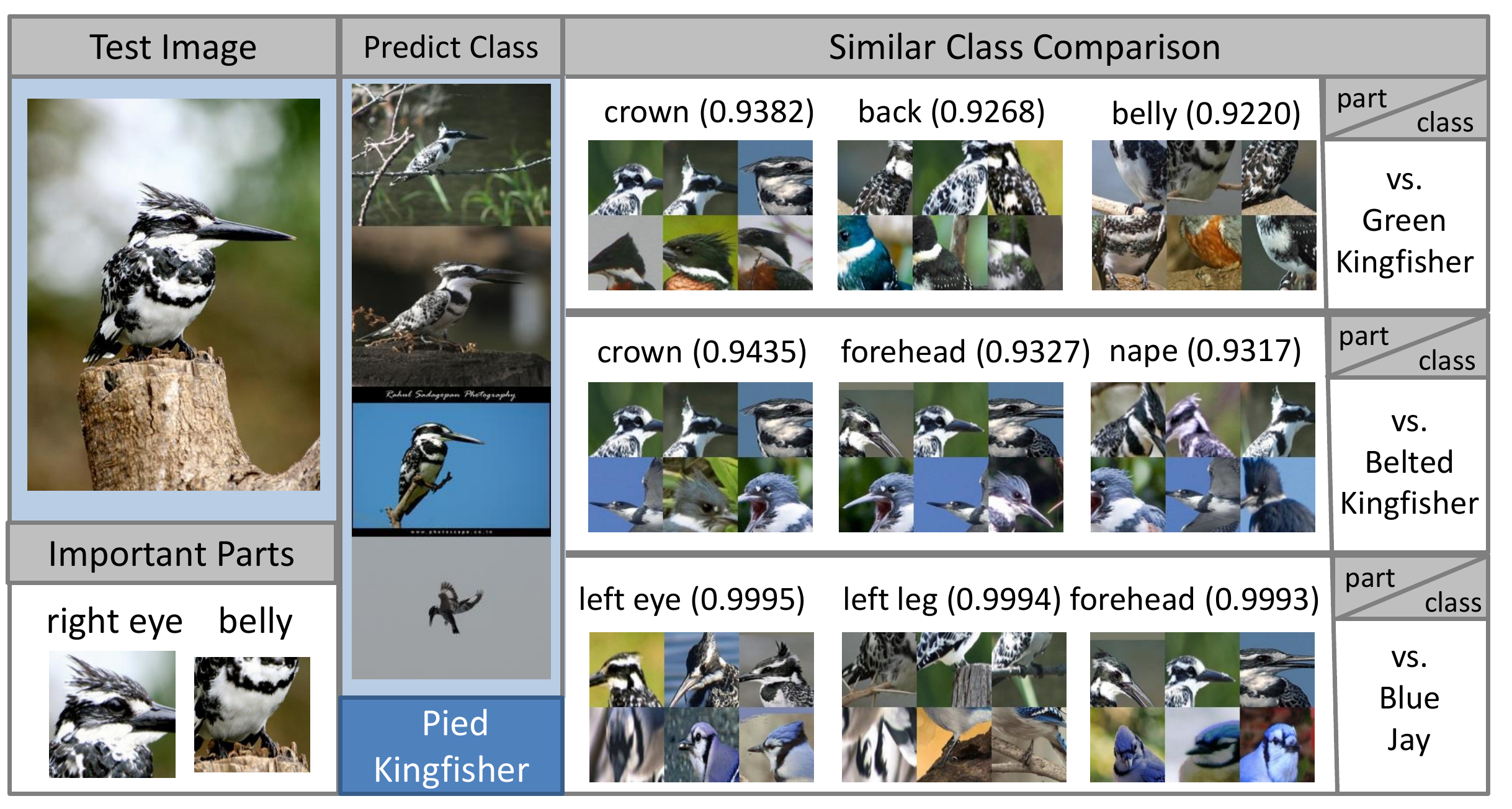}
\end{center}
   \caption{Example of the prediction manual generated by the proposed approach. Given a test image, the system reports its predicted class label with some typical exemplar images. Part-based comparison criteria between the predicted class and its most similar classes are shown in the right part of the image. The number in brackets shows the confidence of classifying two categories by introducing a specific part. We present top three object parts for each pair of comparison. For each of the parts, three part-center-cropped patches are shown for the predicted class (upper rows) and the compared class (lower rows) respectively.}
\label{fig:interpret}
\end{figure*}



We begin the analysis of classification results by a study on the discriminative power of each object part. Each time we select one object part as the input and discard the computation of all other parts. Different parts reveal significantly different classification results. The most discriminative part \emph{crown} itself achieves a quite impressive accuracy of $57\%$, while the lowest accuracy is only $10\%$ for part \emph{beak}. Therefore, to obtain better classification results, it may be beneficial to find a rational combination or order of object parts instead of directly ran the experiments on all parts altogether.


We therefore introduce a strategy that incrementally adds object parts to the whole framework and iteratively trains the model. Specifically, starting from a model trained on bounding-box supervision only, which is also the baseline of the proposed method, we iteratively insert object parts into the framework and re-finetune the PS-CNN model. The number of parts inserted in each iteration increases exponentially, \ie, in the $i^{th}$ iteration, $2^{i}$ parts are selected and inserted. When starting from an initialized model with relatively high performance, introducing a new object part into the framework does not require to run a brand new classification procedure based on this specific part alone; ideally only the classification of highly confusing categories that may be distinguished through the new part will be impacted and amended. As a result, this procedure overcomes the drawback raised by the existence of object parts with lower discriminative power. Table \ref{tab:increm} reveals that as the number of object parts increases from $0$ to $8$, the classification accuracy improves gradually and then becomes saturated. Further increasing the part number does not lead to a better accuracy; however, it does provide more resources for performing explicit model interpretation.

\begin{table}[ht]
\small
\begin{center}
\begin{tabular}{|l|c|c|c|}
\hline
Method                                & Train Anno.& Test Anno. & Acc.\\
\hline
Alignment \cite{gavves2013fine}       & n/a        & n/a        & 53.6 \\
Attention \cite{xiao2014application}  & n/a        & n/a        & 69.7 \\
Bilinear-CNN \cite{lin2015bilinear}   & n/a        & n/a        & 72.5 \\
\hline
CNNaug \cite{razavian2014cnn}         & BBox       & BBox       & 61.8 \\
Alignment \cite{gavves2013fine}       & BBox       & BBox       & 67.0 \\
No parts \cite{krause2015fine}        & BBox       & BBox       & 74.9 \\
Bilinear-CNN \cite{lin2015bilinear}   & BBox       & BBox       & 77.2 \\
\hline
Part R-CNN \cite{zhang2014part}       & BBox+Parts & n/a        & 73.9 \\
PoseNorm CNN \cite{branson2014bird}   & BBox+Parts & n/a        & 75.7 \\
\hline
POOF \cite{berg2013poof}              & BBox+Parts & BBox       & 56.8 \\
DPD+DeCAF\cite{donahue2013decaf}      & BBox+Parts & BBox       & 65.0 \\
Part R-CNN \cite{zhang2014part}       & BBox+Parts & BBox       & 76.4 \\
\textbf{PS-CNN (this paper)}          & \textbf{BBox+Parts} &  \textbf{BBox}       & \textbf{76.2}\\
\hline
\end{tabular}
\end{center}
\caption{Comparison with state-of-the-art methods on the CUB-200-2011 dataset. To conduct fair comparisons, for all the methods using deep features, we report their results on the standard seven-layer architecture (AlexNet) if possible.}
\label{tab:cls}
\end{table}




Table \ref{tab:cls} shows the performance comparison between PS-CNN and existing fine-grained recognition methods. The complete PS-CNN model with a bounding-box and $15$ object parts achieves $76\%$ accuracy, which is comparable with state-of-the-art methods including part-based R-CNN \cite{zhang2014part} and bilinear CNN \cite{lin2015bilinear}. In particular, our model is over two orders of magnitude faster than \cite{zhang2014part}, requiring only $0.05$ seconds to perform end-to-end classification on a test image. This number is quite inspiring, especially considering the number of parts used in the proposed method.

\subsection{Model interpretation}
The proposed approach adopts a part-based strategy to provide visual manuals for fine-grained visual categorization. In particular, we have discovered the most discriminative object parts for classifying a category from other bird species (one-versus-all) and also from its most similar categories (one-versus-most). It is an offline process achieved by calculating the classification performance gain of inserting a part into the bounding-box-only training scheme.

The model interpretation routine is demonstrated in Figure \ref{fig:interpret}. When a test image is presented, the proposed method first conducts object classification through the PS-CNN architecture. The predicted category is presented by a set of images in the dataset that are closest to the test image according to \textit{conv5\_1} outputs. Except for classification results, the proposed method also presents classification criteria for distinguishing the predicted category from its most similar neighbor classes based on object parts. Again we use the output of \textit{conv5\_1} layer but after performing part cropping to retrieve nearest neighbor part patches of the input test image. The procedure described above provides an intuitive visual guide for distinguishing fine-grained categories.

\section{Conclusion}\label{sec:conclusion}
In this paper, we proposed a novel model for fine-grained recognition called Part-Stacked CNN. The model exploited detailed part-level supervision, in which object parts were first located by a fully convolutional network, following by a two-stream classification network that explicitly captured object-level and part-level information. Experiments on the CUB-200-2011 dataset revealed the effectiveness and efficiency of PS-CNN, especially the impact of introducing object parts on fine-grained visual categorization tasks. Meanwhile, we have presented human-understandable interpretations of the proposed method, which can be used as a visual field guide for studying fine-grained categorization.

We have discussed the application of the proposed Part-Stacked CNN on fine-grained visual categorization with strong supervision. In fact, PS-CNN can be easily generalized for varied applications. Examples include:

1) Discarding the requirement of strong supervision. The goal of introducing manually-labeled part annotations here is to generate human-understandable visual guides. However, one can also exploit unsupervised part discover methods \cite{krause2015fine} to define object parts automatically. This strategy should have the potential to achieve comparable classification accuracy with strongly supervised methods, while requiring far less human labeling effort.

2) Attribute learning. The application scenario of PS-CNN is not restricted to fine-grained recognition tasks. For instance, the performance of recommendation systems for online shopping \cite{kiapour2015where} could definitely benefit from the analysis of clothing attributes from local parts. The proposed PS-CNN, in this case, could be an effective and efficient choice to perform part-based garment analysis.

3) Context-based CNN. The role of local ``parts'' in PS-CNN can be also replaced by global contexts. For objects that are small in size and have no obvious object parts, such as volleyballs or tennis balls, modeling global contexts instead of local parts could be a practical solution. Such an architecture can be achieved from the proposed PS-CNN without significant structural changes.

They are regarded as our future works.


{\small
\bibliographystyle{ieee}
\bibliography{egbib}
}

\end{document}